\documentclass[sigconf]{acmart}

\usepackage{array}
\usepackage{times}
\usepackage{soul}
\usepackage{url}
\usepackage[utf8]{inputenc}
\usepackage{caption}
\usepackage{graphicx}
\usepackage{amsmath,amsfonts}
\usepackage{amsthm}
\usepackage{booktabs}
\usepackage{algorithm}
\usepackage{algorithmic}
\usepackage{color}
\usepackage{multirow}
\usepackage{enumitem}
\usepackage{nicefrac}
\usepackage{verbatim}
\usepackage{tablefootnote}

\graphicspath{ {figures/} }

\def\bb{\textbf{b}}
\def\bh{\textbf{h}}
\def\bq{\textbf{q}}
\def\bx{\textbf{x}}
\def\bz{\textbf{z}}
\def\bW{\textbf{W}}

\newcommand{\bM}{{\textbf{M}}}
\newcommand{\mV}{{\mathcal{V}}}
\newcommand{\mE}{{\mathcal{E}}}

\newtheorem{definition}{Definition}

\newtheorem{problem}{Problem}

\acmConference{DLP-KDD 2021}{August 15, 2021}{Singapore}
\acmPrice{}
\acmISBN{}
\acmDOI{}

\newcolumntype{L}[1]{>{\raggedright\let\newline\\\arraybackslash\hspace{0pt}}m{#1}}
\newcolumntype{C}[1]{>{\centering\let\newline  \\\arraybackslash\hspace{0pt}}m{#1}}
\newcolumntype{R}[1]{>{\raggedleft\let\newline \\\arraybackslash\hspace{0pt}}m{#1}}

\begin{document}

\title{TabGNN: Multiplex Graph Neural Network for \\Tabular Data Prediction}


\author{Xiawei Guo\footnotemark[1], Yuhan Quan\footnotemark[2], Huan Zhao\footnotemark[1], Quanming Yao\footnotemark[1]\footnotemark[2], Yong Li\footnotemark[2], Weiwei Tu\footnotemark[1]}
\email{guoxiawei@4paradigm.com; quanyh19@mails.tsinghua.edu.cn}
\email{{zhaohuan, yaoquanming}@4paradigm.com;liyong07@tsinghua.edu.cn; tuweiwei@4paradigm.com}
\affiliation{\institution{\footnotemark[1]4Paradigm Inc., China}
	\institution{\footnotemark[2]Beijing National Research Center for Information Science and Technology(BNRist), \\ Department of Electronic Engineering, Tsinghua University, China}
	    \city{Beijing}
	    \country{China}
}

\newcommand{\huan}[1]{{\color{blue}{\textbf{Huan: }#1}}}
\renewcommand{\shortauthors}{Guo and Quan, et al.}
\newcommand{\huancheck}[1]{{\color{blue}{\textbf{\checkmark\checkmark}#1}}}

\begin{abstract}
  Tabular data prediction (TDP) is one of the most popular industrial applications,
  and various methods have been
  designed to improve the prediction performance.
  However,
  existing works mainly focus on feature interactions and ignore sample relations, e.g.,
  users with the same education level might have a similar ability to repay the debt.
  In this work, by explicitly and systematically modeling sample
  relations, we propose a novel framework TabGNN based on recently popular
  graph neural networks (GNN).
  Specifically, we firstly construct a multiplex graph to model the multifaceted
  sample relations, and then design a multiplex graph neural network to
  learn enhanced representation for each sample.
  To integrate TabGNN with the tabular solution in our company, we concatenate the learned
  embeddings and the original ones, which are then fed to prediction models inside the solution.
  Experiments on eleven TDP datasets from various domains, including classification and regression ones, show that TabGNN can consistently improve the performance compared to the tabular solution AutoFE in 4Paradigm.
  \footnote{The first two authors have equal contributions, and this work is done when Yuhan was an intern in 4Paradigm. Huan Zhao is the corresponding author.} 
\end{abstract}



\maketitle

\section{Introduction}
\begin{figure}[t]
	\centering
	\includegraphics[width=0.95\columnwidth]{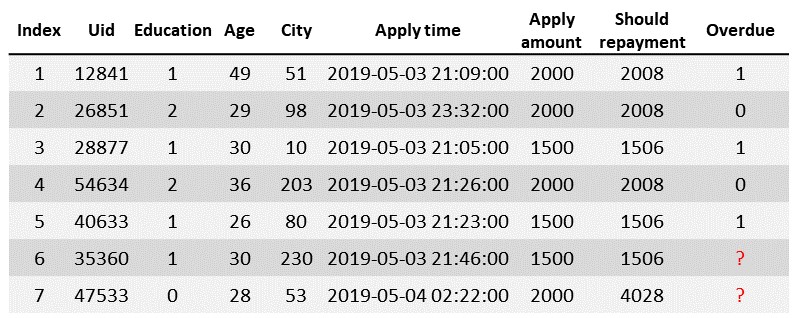}
	\vspace{-10pt}
	\caption{An example tabular data from real-world financial loan activities, where each row represents a sample, and columns are the related features. The values in the column \textit{Overdue} mean whether the user will repay the debt in time. In practice, we need to train a model to predict the values (label) of the column \textit{Overdue} given the information (features) in other columns. Note that the values for \textit{Education} and \textit{City} are encoded as numbers.}
	\label{fig-tabular-example}
\end{figure}

Recent years have witnessed the success of machine learning on various
scenarios, among which tabular data prediction (TDP) is one of the most popular industrial
applications, like fraud detection~\cite{luo2019autocross}, sales prediction~\cite{giering2008retail}, online
advertisement~\cite{rendle2010factorization} and recommender
system~\cite{cheng2016wide,guo2017deepfm}.
In general, data in these cases are in tabular form, where each row represents a
sample and column one feature.
The learning task of TDP
is to predict the values in a target column
given a row (sample) and the corresponding information, e.g., the sales
prediction of a commodity, or whether or not the recorded transaction in a
commercial bank is abnormal.
We give an illustrative example in Figure \ref{fig-tabular-example} to further
explain the concept of this scenario. In practice, TDP is a very
common problem according to a latest
survey from Kaggle~\cite{Kaggle:2020}, the most popular platform for machine
learning competitions. In 4Paradigm, an AI technology and service provider, TDP covers businesses from various customers including commercial banks, hospitals, E-commerce, etc, which contributes to the majority of the revenues. Thus, it is very important to deliver effective TDP solutions.

However, in practice, it is a challenging problem to obtain an effective solution for the TDP task,
since it is far from satisfying to directly use the raw features in tabular forms, i.e., different columns in a row.
To address this challenge, extensive works have been proposed to model feature interactions to improve the prediction performance.
Representative methods include linear models like logistic regression (LR)~\cite{chapelle2014simple}, tree based models like gradient boost machine (GBM)~\cite{friedman2001greedy}, deep models like DeepFM~\cite{guo2017deepfm}, and automated machine learning based models like AutoCross~\cite{luo2019autocross}. Currently, in 4Paradigm,
one of our TDP solutions, \textbf{AutoFE}, includes various modules like automatic feature generation, model tuning and selection, and model serving. AutoFE has been deployed in hundreds of real-world business scenarios from various customers like commercial banks, hospital, media, and E-commerce, etc. For the prediction models, it relies on two popular tabular prediction methods: LR  and GBM.

Despite the fact that these methods including AutoFE have achieved promising results in tabular data applications, they mainly focus on feature interactions, no matter by hand-craft crossing existing features~\cite{chapelle2014simple,friedman2001greedy} or automatically computing higher-order feature interactions~\cite{guo2017deepfm,luo2019autocross}. In this work, we argue that existing methods ignore an important aspect for the TDP task:  the sample relations, which can be very useful for the prediction performance.
We illustrate it by an example in Figure~\ref{fig-tabular-example},
when we want to predict whether user $35360$ ($6$-th row) can repay the debt in time,
those with the same education level, e.g., user $12841$ ($1$st row), $28877$ ($3$rd row), and $40633$ ($5$-th row),
can provide useful signals, since the repayment ability of people with the same education level may be similar.
In manifold learning~\cite{belkin2006manifold} and geometric deep learning~\cite{bronstein2017geometric},
modeling the relations between samples has been
shown beneficial to the empirical performance.
Moreover, the relations among samples can be in multiple facets in tabular data,
leading to a more challenging problem to capture them for the final prediction.
Take the user $35360$ in Figure~\ref{fig-tabular-example} as an example again, those sharing similar ages, i.e., $40633$, $28877$, $35360$, and $47533$, can also provide useful signals. A real-world case in loan default analysis is given in~\cite{hu2020loan}, where the authors show that both the transaction and social relations among users can help predict users' loan default behavior.  While loan default analysis can be naturally modeled as a TDP task, in real-world businesses, the scope of TDP is much more larger, e.g., sales prediction, commodity recommendation, etc. It lacks a solution to capture the sample relations in general tabular scenarios.

In this work, to address this challenge in tabular scenarios, we propose a general TDP solution based on the multiplex graph~\cite{verbrugge1979multiplexity} to capture the multifaceted sample relations. Since multiple edges can simultaneously exist between two nodes in a multiplex graph, it is naturally to represent samples by nodes and relations by edges.
In~\cite{hu2020loan}, an attributed multiplex graph is constructed based on multifaceted relations (social and transfer) to model the loan default behaviors, however, in real-world tabular applications, it is a non-trivial and challenging problem to construct the multiplex graphs, since in many real world tabular applications, the multifaceted sample relations may not explicitly exist. To deal with this problem, we design heuristics to construct the multiplex graph by extracting multifaceted sample relations.
Then we design a multiplex graph neural network to learn a representation of each sample, which is enhanced by neighboring samples from different types of relations. Thus the framework is called TabGNN.
Considering that the feature interaction and sample relations should be complementary for the final prediction, we then integrate TabGNN with the tabular solution AutoFE in 4Paradigm.
To be specific, for each sample, we concatenate the learned embeddings by TabGNN and AutoFE, respectively, and feed them to AutoFE to complete the final prediction.
Experimental results on various real-world industrial datasets demonstrate that TabGNN can further improve the prediction performance compared to AutoFE.
Code is released in Github\footnote{https://github.com/AutoML-4Paradigm/TabGNN.}.

To summarize, the contributions are as follows:
\begin{itemize}[leftmargin=*]
\item Different from most feature interaction based models for tabular data prediction, we propose a complete solution to explicitly and systematically model sample relations in tabular scenarios, which covers various real-world business from our partners.

\item To capture multifaceted sample relations, we propose a novel framework, i.e., TabGNN, which firstly creates multiplex graphs from related samples, and then design a multiplex graph neural network to learn an enhanced representation for each sample.

\item We develop an easy and flexible method to integrate TabGNN with the tabular prediction solution in our company, and extensive experiments on real-world industrial datasets demonstrate that TabGNN can improve the prediction performance compared to the tabular solution in 4Paradigm.
\end{itemize}

\begin{figure*}[t]
	\centering
	\includegraphics[width=0.9\textwidth]{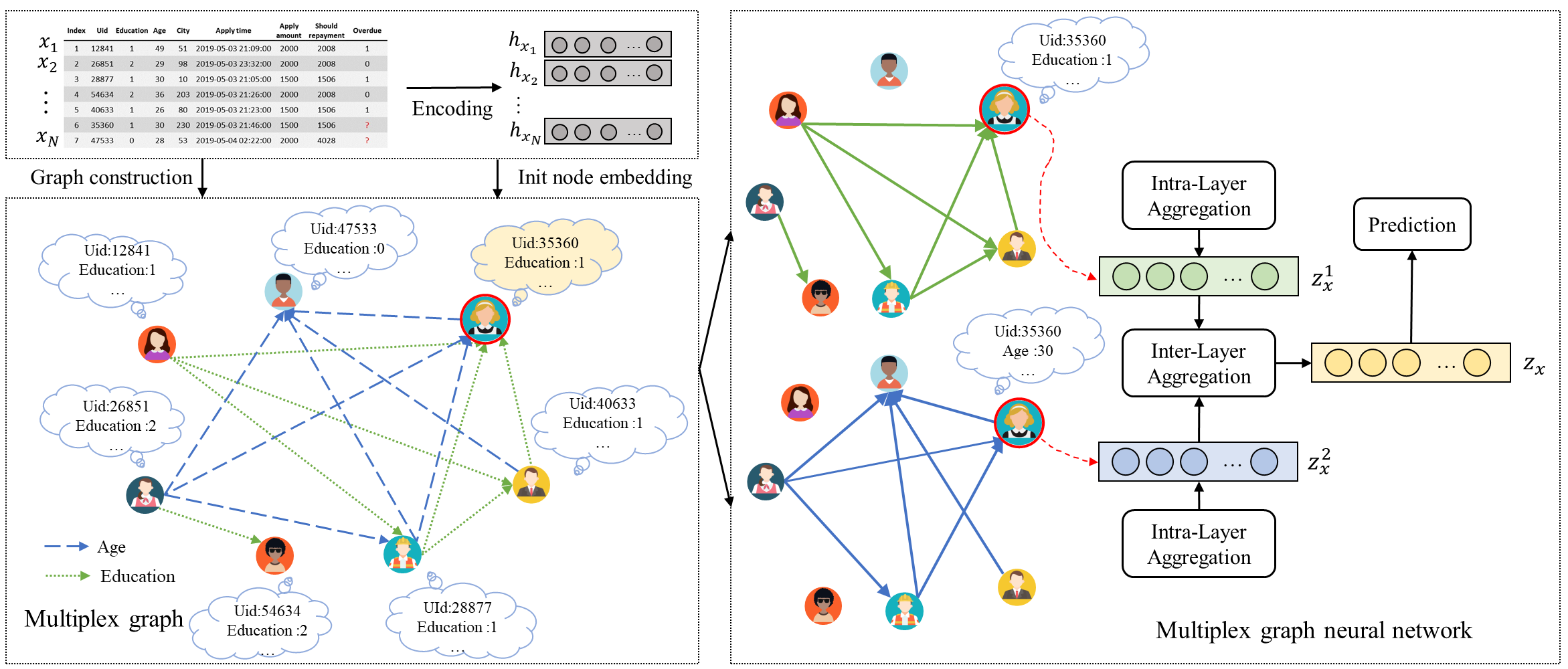}
	\caption{An illustrative example for learning the representations of
            the sample (user $35360$) by TabGNN in Figure~\ref{fig-tabular-example}
            (Best viewed in color).
            Firstly, a directed multiplex graph is constructed based on a
            numerical feature $Age$ and a categorical feature $Education$. The
            features of all samples are encoded to be fed to the multiplex graph neural network to obtain a final representation
            for the sample (user $35360$). A Multilayer Perceptron (MLP) layer
            is used to generate the final prediction label, which is used to
            compute the loss.}
	\label{fig-framework}
	\vspace{-10pt}
\end{figure*}

\section{Related Work}
\label{sec-rel}

\subsection{Existing Methods for  Tabular Data Prediction}

Learning with tabular data is one of the most popular machine learning tasks in industry~\cite{Kaggle:2020}.
To improve the performance of TDP,
various methods were proposed to model feature interactions.
One line of simple methods are (generalized) linear ones, e.g., LR~\cite{chapelle2014simple},
and Wide\&Deep~\cite{cheng2016wide}, which try to feed more hand-crafted features to these models.
Despite the simplicity and good explanation of these methods, they are highly relying on domain knowledge,
thus cannot generalize well in new tasks.
To address this challenge, automated feature crossing based methods,
like DFS~\cite{kanter2015deep}, AutoCross~\cite{luo2019autocross}, AutoFIS~\cite{liu2020autofis}, and AutoFeature~\cite{khawar2020autofeature}, and factorization based methods, like Factorization Machine (FM)~\cite{rendle2010factorization}, DeepFM~\cite{guo2017deepfm},
and PIN~\cite{qu2018product} have been proposed.
The former ones design automated methods to generate cross product of features,
while the latter ones design complex ways to model feature interactions in different orders.
Besides, another line of methods based on gradient boosting~\cite{mason2000boosting} can construct and learn useful features from the raw ones, e.g., GBDT~\cite{friedman2001greedy} and XGBoost~\cite{chen2016xgboost}.

As an AI technology and service provider, we deliver TDP solutions for business customers from various domains in 4Paradigm.
The key solution is AutoFE (Automatic Feature Engine), which includes a whole process for tabular scenarios.
The prediction modules inside AutoFE are relying on LR and GBM, and the features are enhanced by various automatic feature generation methods, mainly relying on feature interactions.
Despite the fact that these methods have achieved promising results in learning with tabular data,
they all focus on feature interactions, while ignore the sample relations.
In this paper, the proposed method try to capture sample relations in tabular data applications,
and thus can be integrated with any feature interaction method for TDP.

\subsection{Multiplex Graph Neural Networks}
\label{sec-rel-multiplex}
Multiplex\footnote{Note that different terms are used in the literature, e.g., multi-dimentional and multi-graph. In this work, we use multiplex for simplicity.} graph~\cite{verbrugge1979multiplexity} was originally designed to model multifaceted relations between peoples in sociology, where multiple edges (relations) can exist between two nodes (people).
Thus, it is natural to use it to model multifaceted relations among two entities, and
has been applied to various applications in recent years, including network embedding~\cite{zhang2018scalable,ma2019multi}, recommendation~\cite{cen2019representation,feng2020mtbrn,zhang2020multiplex}, molecular analysis~\cite{zhang2020molecular}, financial anti-fraud detection~\cite{hu2020loan}, and diagrammatic reason~\cite{Wang2020Abstract}.
Most of these works are trying to model multiple relations on top of the recently popular graph neural networks (GNNs).
Since representative GNN are graph convolutional networks (GCN)~\cite{kipf2016semi}, GraphSAGE~\cite{hamilton2017inductive}, graph attention networks (GAT)~\cite{velivckovic2018graph}, graph isomorphsim network (GIN)~\cite{xu2018powerful}, the above multiplex graph neural networks are built on different GNN models.
In this work, following existing works, we propose to use multiplex graph neural network in a new domain, i.e., tabular data prediction, which is of huge commercial values. Besides, the key difference between our framework and existing works is that they require explicit relations between entities, while in our work, this constraint is not a must. We design various practical heuristics to construct the multiplex graph by extracting explicit and implicit sample relations given a tabular data. Through extensive experiments, we show the effectiveness of multiplex graph construction strategies.

\section{Framework}
\label{sec-framework}

\subsection{Problem Formulation}

For a dataset in tabular form with $N$ rows and $d$ columns, each row
represents a sample $x$, and each column one feature.
For sample $x$, we denote its feature vector by $\bx \in \mathbb{R}^d$.
The label set $\mathcal{Y}$ is given, with $y_x \in \mathcal{Y}$ representing
the label for sample $x$.
In practice, classification and regression are the most two popular tasks in TDP.
For example, it is a classification task to predict whether a user will click an item in a recommendation scenarios, or detect whether a user will repay the debt in a financial scenario, and it is a regression task to predict the sales of a commodity in a store. In this sense, the proposed TabGNN has a broad range of applications in real-world businesses.



For clear presentation, here we formally introduce the multiplex graph following several existing works~\cite{zhang2018scalable,hu2020loan,zhang2020molecular}:
\begin{definition}
	\textbf{Multiplex Graph.}
A multiplex graph can be defined as an $R+1$-tuple $\mathcal{G}=(\mathcal{V},\mathcal{E}^1,\cdots,\mathcal{E}^R)$ where $\mathcal{V}$ is the set of nodes and $\forall r \in \{1,\cdots,R\}, \mathcal{E}^r$ is the set of edges in type $r$ that between pairs of nodes in $\mathcal{V}$. By defining the graph $\mathcal{G}^r = (\mathcal{V}, \mathcal{E}^r)$, which is also called a plex or a layer, the multiplex graph can be seen as the set of graphs $\mathcal{G} = \{\mathcal{G}^1,\cdots,\mathcal{G}^R\}$.
\label{def-multiplex}
\end{definition}

Based on this definition, we then formulate
the problem considered in this work as follow:

\begin{problem}
	\textbf{Multiplex graph based tabular data prediction.}
	Given a tabular dataset, for each sample $x$ with the feature vector $\bx \in \mathbb{R}^d$, a multiplex graph $\mathcal{G}_x=(\mathcal{V}_x,\mathcal{E}_x^1,\cdots,\mathcal{E}_x^R)$ is constructed, where $\mathcal{V}_x$ represents the set of $x$ and the nodes that relate to the $x$ in at least one facet, and $\mathcal{E}_r$ represents the $r$-th type of relation among $\mathcal{V}_x$. Then with $\bx$ and $\mathcal{G}_x$, the problem of multiplex graph based TDP is to predict the label of each sample $x$.
\end{problem}


The overall framework is given in Figure~\ref{fig-framework}.
Given a tabular dataset, we firstly extract several types of sample
relations, based on which a multiplex graph~\cite{verbrugge1979multiplexity} is constructed.
To learn the embeddings of samples, a multiplex graph neural network is designed
to adaptively capture the influences of different neighbors in the multiplex graph.
To train the model,  we use its output  as the predicted label, and
backpropagate the loss to update the parameters.
To integrate the proposed TabGNN with AutoFE, for each sample, we concatenate the learned embeddings by TabGNN and its original features, and feed them to AutoFE again to complete the final prediction.

\subsection{Multiplex Graph Construction}
\label{sec-graph-con}

In this section, we introduce how to construct multiplex graphs by capturing the multifaceted sample relations in tabular scenarios.
Practically speaking, there are two challenges in constructing the multiplex graph: the relation extraction and temporal constraint.

\vspace{3pt}
\noindent\textbf{Relation extraction.} The first challenge is to extract multifaceted relations among samples.
Different from existing multiplex graph based works in Section \ref{sec-rel-multiplex}, where the multiplex graphs are given directly, in tabular scenarios, sample relations are not always given. Therefore, we need to extract the relations from data.
In practice, tabular data are from highly diverse domains,
e.g., fraud detection, or recommender systems,
resulting in various types of relations that can be inferred by their features.
As a result, we need to construct the graphs by features.
In this work, based on our industrial experiences, we give some heuristics in the following:
\begin{itemize}[leftmargin=*]
\item The first one is to consider to connect samples that have the same value in
some  ID features.
For example, in a click-through rate (CTR)
prediction scenario, samples with the same user ID might be strongly related~\cite{rendle2010factorization}, and therefore should be connected in the graph.


\item Beyond ID features, we can choose categorical features of highly
    important scores, which can be obtained by feature selection methods,
    e.g., AutoFE.
For example, in a financial loan scenario (Figure~\ref{fig-tabular-example}),
samples (users) with the same \textit{Education} level may show a similar
ability to pay the debt.

\item Numerical features
    are usually discretized in learning tasks
    to generate categorical features~\cite{liu2002discretization},
    which can also be selected to construct the graph.
Again in the same example in Figure~\ref{fig-tabular-example}, samples (users)
with similar \textit{Age}, e.g., $[25,30]$, may show similar ability to pay
the debt.

\item Besides, we can take the product of categorical features. For example,
    samples (users) with same \textit{Education} level and similar
    \textit{Age} interval
may show similar ability to pay the debt in Figure~\ref{fig-tabular-example},
which should thus be connected.
\end{itemize}
Note that our purpose is not to exhaustively enumerate heuristics in all
real-world scenarios, which should be an impossible task, but to show some
common ones according to our practical experiences from various domains.
They can thus be used as a quick start to apply TabGNN to a new domain.

\vspace{3pt}
\noindent\textbf{Temporal constraint.}
The second challenge is to deal with the \textit{temporal constraint},
which is that the preceding samples cannot ``see'' the information, e.g., labels, of the subsequent samples. Considering that samples tend to be attached with timestamps, e.g., Apply time in Figure \ref{fig-tabular-example}, it can lead to data leakage problem~\cite{kaufman2011leakage} if the temporal constraint is not processed properly.
%
To address this challenge, we use directed edges when constructing the graphs, where the
preceding samples are pointing to the subsequent ones,
to avoid a sample to``see'' its ``future'' samples, as shown in
Figure~\ref{fig-framework}.
Note that there can be tabular data without temporal constraint (``Data3'' used in our experiments), although it might be not that common, we can
create undirected multiplex graphs correspondingly.


\subsection{Multiplex Graph Neural Network}
\label{subsec-learn-gnn}
In this section, we design a multiplex graph neural network to learn the representation of each sample, consisting of four steps: feature encoding,
intra-layer aggregation, inter-layer aggregation,
and model training, as shown in the bottom part of Figure \ref{fig-framework}.

In Definition \ref{def-multiplex},  a multiplex graph $\mathcal{G} = (\mathcal{V}, \mathcal{E}^1, \cdots, \mathcal{E}^L)$ can be seen as a set of graphs $\{\mathcal{G}^1,\cdots,\mathcal{G}^R\}$, where $\mathcal{G}_r$ represents the $r$-th type of relation between nodes in $\mathcal{V}$ and called a layer. Thus, for each $\mathcal{G}_r$, we design a GNN model to update the embedding of the sample $x$ by neighbors from the $r$th type of relation. For each sample $x$, we can obtain $R$ groups of embeddings based on $R$ GNN models, each of which contributes differently to the final embeddings of $x$, then we design an attention mechanism to aggregate these $R$ groups of embeddings to learn the final embeddings of $x$. Since each graph $\mathcal{G}_r$ is called plex or layer and the key module of GNN is a neighborhood aggregation function, we denote these two parts by \textit{Intra-Layer Aggregation} and \textit{Inter-Layer Aggregation}, respectively.

\vspace{2pt}
\noindent
\textbf{Feature Encoding.}
Firstly, for a sample $x$ in tabular data,
its associated features,
i.e., the
columns in the corresponding row in the table,
are usually of different types, e.g.,
the salary of a user is numerical, while the
gender of a user is categorical.
Therefore, the feature vector $\bx$ cannot be directly fed to GNN, and we design feature encoders to transform
different types of features into a unifying latent space, which is denoted as
$\bh_x = ENC(\bx)$, where $ENC(\cdot)$ represents the feature-wise encoder.
Details about encoders are given in Appendix D.



\vspace{2pt}
\noindent\textbf{Intra-Layer Aggregation.}
Given a sample $x$ with its encoded feature vector $\bh_x$, before the intra-layer aggregation, we firstly transform the $\bh_x$ into a new space by a projection matrix $\bM^r$. The underlying assumption is that different layers should have different latent feature space, for which we use a relation-specific weighting matrix $\bM^r$. Then, the transformed feature of the sample $x$ is $\hat{\bh}_x = \bM^r \cdot \bh_x$.
Then for the $r$-th layer, the embedding of sample $x$, $\bz_x^{r} \in \mathbb{R}^{d_1}$,  is computed as the following:
\begin{equation}
\label{eq-mpnn}
\bz_x^{r} =
\sigma
\left(
\bW^{r} \cdot \text{AGG}(\{\hat{\bh}_u, \forall u \in \widetilde{N}^{r}(x)\})
\right),
\end{equation}
where $\bW^{r} \in \mathbb{R}^{d_1 \times d_0}$ is a trainable weight matrix
shared by all samples, and
$\sigma$ is a non-linear activation function, e.g., a sigmoid or ReLU.
$\text{AGG}$ is the aggregation function, corresponding to different GNN models, like GCN~\cite{kipf2016semi} or GAT~\cite{velivckovic2018graph}.
In our implementations,we choose the attention aggregation function, which is widely used in GAT.
In the experiments, we conduct ablation studies on different aggregation functions.
$\widetilde{N}^{r}(x) = \{x\} \cup N^{r}(x)$,
where $N^{r}(x)$ is the incoming neighbor set of node $x$ in the $r$-th layer.

\begin{algorithm}[t]
	\caption{The whole process of TabGNN.}
	\label{alg-tabgnn}
	\begin{algorithmic}[1]
		\REQUIRE The multiplex graph
		$\{(\mV, \mE^1, \mE^2, \cdots, \mE^R)\}$ \\
		The node features $\{\bx \in \mathbb{R}^{d}| x \in \mV\}$ \\
		\ENSURE  The of Intra-Layer Aggregation including
		$\{\bM^r, \bW^r | r = 1, 2,\ldots, R\}$ and AGG. \\
		The parameters of the Inter-layer Aggregation $\bq, \bW, \bb$.\\
		The parameters of predictors $\bW_o, \bb_o$.
		\STATE  Encode features $\bh_x = ENC(\bx)$  for all $x \in \mV$.
		\FOR{$r = 1, 2, \cdots, R$}
		\FOR{$x \in \mV$}
		\STATE Find $\tilde{N}^r(x)$ according to $\mE_r$.
		\STATE  Transform the encoded features $\hat{\bh}_x = \bM^r \cdot \bh_x$.
		\STATE Update the embedding $\bz_x^{r}$ with $\tilde{N}^r(x)$ using \eqref{eq-mpnn}.
		\ENDFOR
		\ENDFOR
		\FOR{$x \in \mV$}
		\STATE Obtain the embedding $\bz_x$ using \eqref{eq-s}, \eqref{eq-r}, and \eqref{eq-fuse}.
		\ENDFOR
		\STATE Calculate the loss of $\bz_x$ by \eqref{eq-loss}.
		\STATE Back propagate the loss and update the parameters.
	\end{algorithmic}
\end{algorithm}

\vspace{2pt}
\noindent\textbf{Inter-Layer Aggregation.}
To adaptively fuse the $R$ groups of representations, we then design a
one-layer neural network as following:
\begin{align}
s^r
& = \nicefrac{1}{|\mathcal{V}|} \sum\nolimits_{x \in \mathcal{V}} \bq^{\top}
\!\!
\cdot \text{tanh}(\bW \cdot \bz^r_x + \bb),
\label{eq-s}
\\
\beta^r
& = \exp(s^r) / \sum\nolimits_{r'=1}^{R}\exp(s^{r'}),
\label{eq-r}
\\
\bz_x
& = \sum\nolimits_{r=1}^{R} \beta^r \cdot \bz_x^r.
\label{eq-fuse}
\end{align}

\vspace{2pt}
\noindent\textbf{Model Training.}
Then $\bz_x$ is used as the final representation of the sample $x$
to learn a classification or regression function on the train data by the following framework:
\begin{equation} \label{eq-loss}
\sum\nolimits_{x \in \mathcal{V}}
l
\big( y_x, \sigma(\bW_o^T\bz_x + \bb_o) \big),
\end{equation}
where $l(\cdot)$ is the cross-entropy or mean square loss according to the specific problem, and $\bW_o, \bb_0$ are the parameters. The whole process of TabGNN is given in Algorithm \ref{alg-tabgnn}.

\subsection{Integration with AutoFE}
\label{sec-framework-integration}
It is intuitive that both feature interaction and sample relations can be
useful for TDP tasks, we then integrate TabGNN with the tabular solution in
4Paradigm. Here we briefly introduce AutoFE and then how we integrate
TabGNN with it.
\subsubsection{AutoFE}
\label{sec:hyml}
\begin{figure}[t]
	\centering
	\includegraphics[width=0.85\columnwidth]{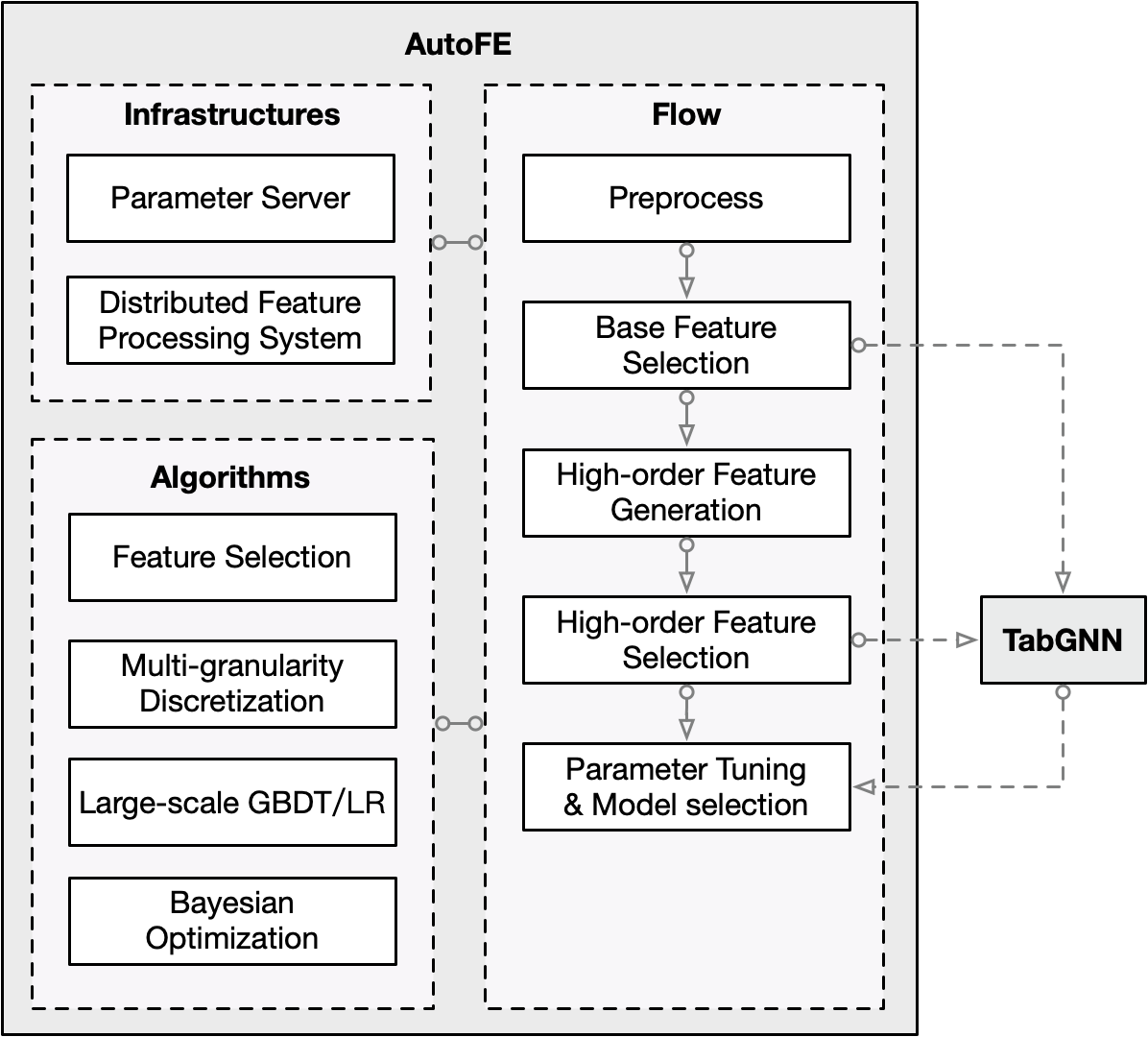}
        \caption{The system overview of AutoFE. We further add TabGNN to show the integration process.}
        \vspace{-5pt}
	\label{fig:hcml}
\end{figure}
To handle large scale machine learning problems,
AutoFE is implemented based on distributed computing infrastructures, which can
run on a single machine or a cluster.
It can automatically generate and select important high-order features.
The architecture of AutoFE is shown in the ``Flow'' part of
Figure~\ref{fig:hcml}.
The key steps are explained as follows.
\begin{enumerate}
    \item \textbf{Preprocess}
The input data will first be preprocessed, where noisy
rows or columns are dropped.  We denote the valid features of the
input data as base features.
    \item \textbf{Base Feature Selection}
Importance scores are calculated for base features by a feature
selection process, where a LR model is trained on each
feature in parallel to test their performance on validation set.
    \item\textbf{High-order Feature Generation}
The system then generated high-order features by combining base features.
AutoFE implements a set of feature combination operators and assigns
an importance score to each of them. A high-order feature will be generated
if the sum of importance scores over its two base features and the combination
operator is the $K$th largest, where $K$ is predefined.
A distributed feature processing module is used to handle the
large scale feature combination problem.
    \item \textbf{High-order Feature Selection}
A subset of high-order features are further selected by
testing their contribution to performance with
all base features on the validation dataset.
This subset of features are finally concatenated with the base features to
generate AutoFE feature embeddings.
\item \textbf{Parameter Tuning \& Model Selection}
Based on the feature embeddings, Bayesian Optimization is used to select
models and tune their hyper-parameters.
AutoFE implements models including large scale GBM and
LR based on a parameter server.  For LR,
multi-granularity discretization is used to enhance the representation ability
for numerical features.
\end{enumerate}

\subsubsection{Integration}
The prediction models of AutoFE, the tabular solution in our company,
are two widely used models: LR and
GBM. Note that unlike those technology giant companies,
like Google or Alibaba, where huge neural network models are more commonly
used, most of our customers are from more traditional businesses, like bank or
E-commerce, where LR and GBM are more widely used. It also aligns with a latest
report from Kaggle~\cite{Kaggle:2020} in 2020, that LR and GBM are the most
popular models in data science.

Therefore, in this work, to integrate the proposed TabGNN with our tabular solution,
we adopt a straightforward method.
The multiplex graph is constructed based on the base features with their importance
scores, and the embeddings of each sample output by TabGNN is used as extra
features fed to AutoFE in the model seletion and parameter tuning step.
To be specific, for sample $x$ with its AutoFE embedding $\bx$, we obtain a representation $\bz_x$ by TabGNN,
and then concatenate $\bx$ and $\bz_x$ as features.
It is an easy and flexible manner to deploy the proposed TabGNN in real-world business,
and this simple pipeline has also adopted by previous works~\cite{rendle2010factorization,covington2016deep,grbovic2018real}.
In the experiment, besides AutoFE,
we further inject the learned representation into a popular deep neural network model,
i.e., DeepFM \cite{guo2017deepfm}, to show the usefulness of TabGNN with more complex models.

\subsection{Discussion}

Since one of the most popular TDP scenarios is the click-through-rate (CTR) prediction, where each sample is a user-item pair representing the clicked behavior of the user. In this scenario, sample relations exist inherently, because samples can be related by the same user clicking different items or the same item clicked by different users. In the literature, various models have been proposed to capture the sample relations in CTR scenarios, like Deep Interest Network (DIN) \cite{zhou2018deep} and Behavior Sequence Transformer (BST) \cite{chen2019behavior}, which captures the most popular sample relation, i.e., the sequential one, due to the sequential nature of users' clicked items.

\begin{table*}[t]
	\setlength\tabcolsep{5pt}
	\centering
	\caption{Statistics of datasets. ``\#Num'' and ``\#Cat'' represents the number of numerical and categorical features, respectively. ``N'' means no temporal constraint for Data3.}
	\begin{tabular}{c|ccccccccc}
		\toprule
		& \multirow{2}{*}{Task}& \multirow{2}{*}{Dataset}& \multicolumn{2}{c}{\#Samples} & \multicolumn{2}{c}{\#Features} & Validation &\multirow{2}{*}{Domain} &  \multirow{2}{30pt}{Temporal constraint} \\
		& && Train     & Test      & \#Num  & \#Cat & Ratio\\ \midrule
		\multirow{9}{*}{Private}
		& Classification& Data1        & 35,581    & 8,895     & 16   & 17  & 15\% & Loan           & Y \\
		& Classification& Data2        & 1,888,366 & 1,119,778 & 8    & 23  & 5\%  & News           & Y \\
		& Classification& Data3        & 108,801   & 27,201    & 19   & 9   & 15\% & Loan           & N \\
		&   Classification & Data4        & 226,091   & 34,867    & 14   & 26  & 10\% & E-commerce     & Y \\
		&   Classification & Data5        & 435,329   & 31,076    & 8    & 34  & 10\% & E-commerce     & Y \\\cline{2-10}
		&   Regression & Data6        & 1,638,193 & 702,016   & 43   & 16  & 10\% & Live streaming & Y \\
		&   Regression & Data7        & 3,923,406 & 694,194   & 0    & 25  & 5\%  & Retail         & Y \\
		& Regression& Data8        & 10,512,133& 29879     & 4    & 17  & 5\%  & Retail         & Y \\
		& Regression& Data9        & 179,893    & 43,236     & 5    & 2   & 15\% & Government       & Y \\
		\midrule
		\multirow{2}{*}{Public}
		& Classification & Home Credit  & 307,511   & 48,744    & 175  & 51  & 10\% & Loan            & Y \\
		& Classification & JD           & 4,992,910 & 446,763   & 6    & 17  & 5\%  & E-commerce      & Y \\
		\bottomrule
	\end{tabular}
	\label{table-dataset}
\end{table*}

Despite the similarity in modeling sample relations between sequential CTR models and TabGNN, here, we try to clarify the differences.
The first difference is that the sequential CTR models tend to require the explicit existence and property (sequential) of the sample relations, e.g., by user IDs, while in TabGNN, we do not have this requirement. We can model both explicit and implicit sample relations by constructing the multiplex graphs.
The second difference is that the scope of TDP is much larger than CTR scenarios, like sales prediction and fraud detection. Therefore, the popular sequential CTR models can be seen as a special case of TabGNN.

\section{Experiments}
\label{sec-exp}

In this section, we conduct experiments in our business partners from various domains to show the effectiveness of TabGNN compared with the tabular solution AutoFE in 4Paradigm.

\subsection{Experimental Settings}

\noindent\textbf{Tasks and Datasets.}
For the tasks, we choose both classification and regression tasks, and the details of datasets are shown in table~\ref{table-dataset}.

To be specific, we use 9 private datasets from 6 different domains, which are provided by our business customers.
Regarding the specific tasks in these domains, firstly, for the classification task, "Loan" means predicting whether a user can repay the loan on time, "E-commerce" means predicting whether the user will buy a product on the website, and "News" means predicting whether the user will click a news on the website. For regression tasks, "Live-streaming" means predicting the length of time users will watch a live broadcast, "Retail" means predicting the sales of goods in the store, and "Government" means predicting the public's attention(number of comments, suggestions) to some public issues in the city.

We further use two public tabular datasets, Home Credit and JD. Home Credit\footnote{https://www.kaggle.com/c/home-credit-default-risk/overview} is a well-known tabular dataset in Kaggle, which aims to predict clients' repayment abilities. JD dataset is extracted from a competition\footnote{https://jdata.jd.com/html/detail.html?id=1} hosted by JD Inc.,
whose task is to predict whether users will purchase products from a given list, we turn it into a standard binary classification task, we explained in detail in Appendix A. Note that there are multiple tables in JD and Home Credit, an module in AutoFE can transform them into single tables through some database operations, which concatenate features from different tables based on some keys for the same samples. All prediction methods are then executed based on the transformed single tables.

\vspace{2pt}
\noindent\textbf{Evaluation metrics.} For the classification task, we use AUC as the evaluation metric, and for the regression task, we use the mean square error (MSR) as the evaluation metric. For AUC, large values means better performance, while for MSE, smaller values means better performance.

\vspace{2pt}
\noindent\textbf{Implementation Details.}
To show the effectiveness of TabGNN, we integrate it with the tabular solution AutoFE in our company, whose prediction models are relying on two popular tabular models in practice: LR and GBM. Note that AutoFE can automatically select the best model from LR and GBM based on some built-in model selection strategy. We then report the performance by AutoFE with and without the embeddings from TabGNN.

 For the implementation of TabGNN, we use a popular open source library for GNNs: DGL\footnote{https://github.com/dmlc/dgl} (Version 0.5.2) and PyTorch (Version 1.5.0), and build our code based on an existing codebase Table2Graph \footnote{https://github.com/mwcvitkovic/Supervised-Learning-on-Relational-Databases-with-GNNs}. Other environments includes: Python 3.7, Linux (CentOS release 7.7.1908) server with Intel Xeon Silver 4214@2.20GHz, 512G RAM and 8 NVIDIA GeForce RTX 2080TI-11GB.

For DeepFM, we use an open source implementation\footnote{https://github.com/shenweichen/DeepCTR-Torch}.


\begin{table*}[t]
	\setlength\tabcolsep{6pt}
	\centering
	\caption{The results of different datasets. For the classification and regression task, we use AUC and MSE as the evaluation metrics, respectively. We further show the improvements compared to baselines.}
	\begin{tabular}{c|ccccccccc|cc}
		\toprule
			& \multicolumn{9}{c|}{Private} & \multicolumn{2}{c}{Public} \\\midrule
		Task & \multicolumn{5}{c|}{Classification (AUC)} & \multicolumn{4}{c|}{Regression (MSE)} & \multicolumn{2}{c}{Classification (AUC)} \\
			\midrule
		Dataset & Data1 &  Data2  &  Data3  & Data4 & \multicolumn{1}{c|}{Data5} &Data6 & Data7 & Data8 & Data9 & Home Credit & JD \\
		\midrule
		AutoFE                & 0.6021 & 0.8662 & 0.9019 & 0.9787 & \multicolumn{1}{c|}{0.8310} & 15726.31& 10.94 & 20.47 & 196.47 & 0.7298 & 0.7159 \\
		+TabGNN         & \textbf{0.6139} & \textbf{0.8929} & \textbf{0.9139} & \textbf{0.9857} & \multicolumn{1}{c|}{\textbf{0.8754}} & \textbf{14392.52}& \textbf{10.05} & \textbf{19.93} & \textbf{191.24} & \textbf{0.7408} & \textbf{0.7537}\\
		Improvement    &  1.9\% & 3.1\% & 1.3\% & 0.7\% & \multicolumn{1}{c|}{5.3\%} & 5.8\%& 8.1\%& 2.6\% & 2.7\% & 1.5\% & 5.3\% \\
				\midrule
		DeepFM            & 0.5953 & 0.8887 & 0.8878 & 0.9838 & \multicolumn{1}{c|}{0.8151} & 14377.51 & 15.44 & 21.98 & 197.10 & 0.7028  & 0.7021 \\
		+TabGNN     & \textbf{0.5972} & \textbf{0.9062} & \textbf{0.9141} & \textbf{0.9846} & \multicolumn{1}{c|}{\textbf{0.8411}} & \textbf{14306.03} & \textbf{11.07} & \textbf{20.51} & \textbf{191.71} & \textbf{0.7316} & \textbf{0.7536}\\
		Improvement    & 0.3\%  & 2.0\% & 3.0\% & 0.1\% & \multicolumn{1}{c|}{3.2\%} & 0.5\% & 28.3\% & 6.7\% & 2.7\% & 4.1\% & 7.3\% \\
		\bottomrule
	\end{tabular}
	\label{table-performance}
\end{table*}



\vspace{2pt}
\noindent\textbf{Multiplex Graph Construction.}
Here we introduce the extracted relations to construct the multiplex graph for each dataset. The details are given in Table \ref{table-dataset-graph-construction}, in which we briefly explain the features used as relations.

\vspace{2pt}
\noindent\textbf{Training Details.}
\label{Parameter-tuning}
The training and test datasets are given in Table \ref{table-dataset}. And for each dataset, we choose samples ranging from 5\% to 15\% as the validation set according to the size of the training data, and the validation ratio is also given in Table \ref{table-dataset}. 
For datasets with temporal constraint, we select the validation samples by the latest timestamps.
For the datasets without temporal constraint, we randomly choose the validation samples. For hyper-parameter tuning, we refer readers to Appendix E.

\subsection{Experimental Results}

The results are given in Table \ref{table-performance}, from which we can see that it can consistently improve the prediction performance significantly by integrating TabGNN with AutoFE and DeepFM on both the classification and regression tasks.
Considering the variety of these 11 datasets in terms of TDP scenarios, it demonstrates the effectiveness of the proposed framework in broader domains. In other words, it shows the usefulness of the sample relations for TDP.
Besides, by the straightforward method to integrate TabGNN with AutoFE in Section \ref{sec-framework-integration}, it means that this framework can be quickly adopted in new business scenarios where AutoFE has been deployed, which can bring huge commercial values to both consumers and 4Paradigm.

Besides, an interesting observation is that AutoFE, whose prediction modules are either LR or GBDT,  outperforms DeepFM in most cases.
We attribute this to the engineering pipeline, especially the feature selection, model selection and tunning, which is continually optimized by practical experiences from hundreds of TDP scenarios from our business customers.
This phenomenon gives two further implications. The first one is that shallow models can beat deep models in real-world businesses with effective engineering strategies. Secondly, since TabGNN can improve the performance compared with both shallow and deep models, it means sample relations are actually not well modeled by previous feature interaction methods for TDP.

\begin{table}[ht]
	\setlength\tabcolsep{5pt}
	\centering
	\caption{Construction details of the multiplex graphs. The Condition column represents two samples (nodes) are connected when the condition is satisfied.}
	\begin{tabular}{p{30pt}<{\centering}p{100pt}<{\centering}p{80pt}<{\centering}}
		\toprule
		Dataset & Features & Condition \\
		\midrule
		\multirow{2}{*}{Data1} & Product of 4 features\tablefootnote{\textit{sex}, \textit{profession}, \textit{education}, \textit{marriage}} & same  \\
		& \textit{browse\_history}\tablefootnote{Construct a vector for each user based on its browse history. If the user browses the corresponding content, the value of this element is 1, otherwise it is 0. } & TopK similar users \\
		\midrule
		\multirow{2}{*}{Data2} & \textit{user\_id} & Same  \\
		& \textit{new\_id}  & Same  \\
		\midrule
		\multirow{2}{*}{Data3} & \textit{age}  & Difference $\leq$ 2  \\
		& \textit{city} & Same   \\
		\midrule
		\multirow{2}{*}{Data4} & \textit{user\_id} & Same  \\
		& \textit{item\_id} & Same  \\
		\midrule
		\multirow{2}{*}{Data5} & \textit{user\_id} & Same  \\
		& \textit{city}     & Same  \\
		\midrule
		\multirow{2}{*}{Data6} & \textit{user\_id} & Same  \\
		& \textit{host\_id}     & Same  \\
		\midrule
		\multirow{3}{*}{Data7} & \textit{store\_id} & Same  \\
		& \textit{item\_id}     & Same  \\
		& \textit{distinct\_id}     & Same  \\
		\midrule
		\multirow{2}{*}{Data8} & \textit{item\_id} & Same  \\
		& \textit{brand\_id}     & Same  \\
		\midrule
		\multirow{2}{*}{Data9} & \textit{location} & K closest  \\
		& \textit{event\_type}     &  Same \\
		\midrule
		\multirow{2}{*}{Home Credit} & \textit{income} & Difference $\leq$ 2000  \\
		& Product of 6 features\tablefootnote{\textit{CODE\_GENDER}, \textit{FLAG\_OWN\_CAR}, \textit{FLAG\_OWN\_REALTY}, \textit{FLAG\_MOBIL}, \textit{FLAG\_EMP\_PHONE}, \textit{FLAG\_CONT\_MOBILE}} & Same \\
		\midrule
		\multirow{2}{*}{JD} & \textit{user\_id} & Same  \\
		& \textit{sku\_id}  & Same  \\
		\bottomrule
	\end{tabular}
	\label{table-dataset-graph-construction}
\end{table}

Finally, we emphasize one more observation that the performance gains in Data7 are the largest among all datasets. It is very likely to be attributed to that we construct the multiplex graph with three types of relations while on other datasets only two type of relations are used. Therefore, taking all these observations into consideration, we can empirically verify the usefulness of sample relations for TDP.


%

\subsection{Ablation Study}
\label{append-ablation}
In this section, we conduct ablation studies on the proposed TabGNN,
including different GNN variants and graph construction methods. For simplicity, we use two classification datasets, Data3 and Home Credit, and one regression dataset: Data6.

\begin{table}[ht]
	\setlength\tabcolsep{7pt}
	\centering
	\caption{The performance of different GNN variants. Note that AUC and MSE are the evaluation metrics for classification and regression tasks, respectively.}
	\begin{tabular}{cccc}
		\toprule
		Task & \multicolumn{2}{c}{Classification (AUC)} & Regression (MSE)\\
		Dataset & Data3   & Home Credit & Data6\\
		\midrule
		GCN       & 0.8911 & 0.7372 & 14563.4 \\
		GAT       & 0.8965 & 0.7387 & 14404.2 \\
		\midrule
		TabGNN    & 0.9139 & 0.7408 & 14392.5 \\
		\bottomrule
	\end{tabular}
	\label{table-gnn-variant}
\end{table}

\subsubsection{GNN variants}
\label{append-ablation-gnn}
The proposed TabGNN is a multiplex GNN.
Here, we show the performance comparisons between TabGNN and two
popular GNN models for homogeneous graphs:
GCN~\cite{kipf2016semi}, GAT~\cite{velivckovic2018graph}.
Since GCN and GAT are designed for
homogeneous graph, i.e., edges are of no types,
we reduce the constructed graph to homogeneous one by merging multiple edges between two nodes as one, and apply these two models.
The results are shown in Table~\ref{table-gnn-variant}, from which we can see
that TabGNN outperforms GCN and GAT consistently on Data3, Home Credit and Data6. It demonstrates the effectiveness of the proposed multiplex graph neural network in Section \ref{subsec-learn-gnn}.

\begin{table}[h]
	\setlength\tabcolsep{7pt}
	\centering
	\caption{The results of graphs constructed by different relations. Note that AUC and MSE are the evaluation metrics for classification and regression tasks, respectively.}
	\begin{tabular}{ccccc}
		\toprule
		Task & \multicolumn{2}{c}{Classification (AUC)} & Regression (MSE) \\
		Dataset & Data3   & Home Credit & Data6\\
		\midrule
		Relation 1      & 0.8975 & 0.7388 & 14434.8 \\
		Relation 2      & 0.8979 & 0.7394 & 14411.9 \\\midrule
		Both           & 0.9139 & 0.7408 & 14392.5 \\
		\bottomrule
	\end{tabular}
	\label{table-graph-construction}
\end{table}

\subsubsection{Influence of number of relations}
\label{append-ablation-correlation}
As introduced in Section \ref{sec-graph-con}, we construct a multiplex graph
to capture multifaceted sample relations. Here we show the influence of
the number of relation types in the constructed graph. For these three datasets,
we design two types of relations to construct the graph as shown in Table \ref{table-dataset-graph-construction}, and then we show the performance comparisons between a single type of
relations and both ones. The results are shown in Table
\ref{table-graph-construction}.  We can see that the performance gains are
obvious by incorporating both relations, which implies the
effectiveness of multiple facets of the sample relations.

\begin{figure}[ht]
	\centering
	\includegraphics[width=0.7\columnwidth]{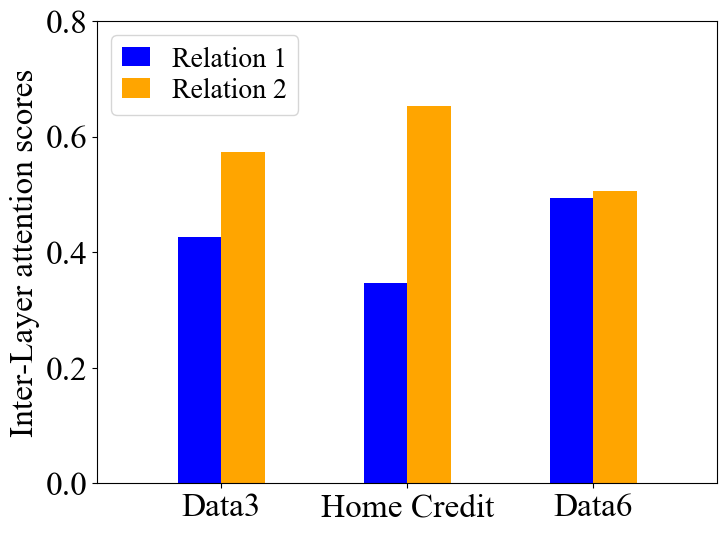}
	\caption{Attention scores of different relations by TabGNN.}
	\label{fig-attention-score}
\end{figure}
In Section \ref{subsec-learn-gnn}, we have a Inter-Layer Aggregation module to model different contributions of each relation to the final predictions. Here in Figure~\ref{fig-attention-score}, we further visualize the attention scores of different relations by TabGNN on these three datasets.
We can see on all datasets relation 2 is larger than relation 1,
which is consistent to the result we obtained in Table~\ref{table-graph-construction}
where only using relation 2 performs better than relation 1.

Moreover, we give a in-depth analysis on Data3, which is a dataset
from loan activities, and the two relations are from \textit{Age} and
\textit{City}. The higher attention score of relation \textit{City}
implies a higher utilization of information from other people in the same city
in predicting a person's repayment ability.
Similarly on Home Credit, the first relation is
from feature \textit{Income}, and the second relation is from a combination
of features related to users' assets like cars or
realities. Thus, the larger attention score of the second relation
shows stronger signals from users with similar assert states
when evaluating the repayment ability of a user.
For Data6, it is a live streaming data, where \textit{user\_id} and \textit{host\_id} are used to construct the multiplex graphs, and we can see that the importance of these two relations are similar for the final prediction.

\begin{figure}[ht]
	\centering
	\includegraphics[width=0.85\columnwidth]{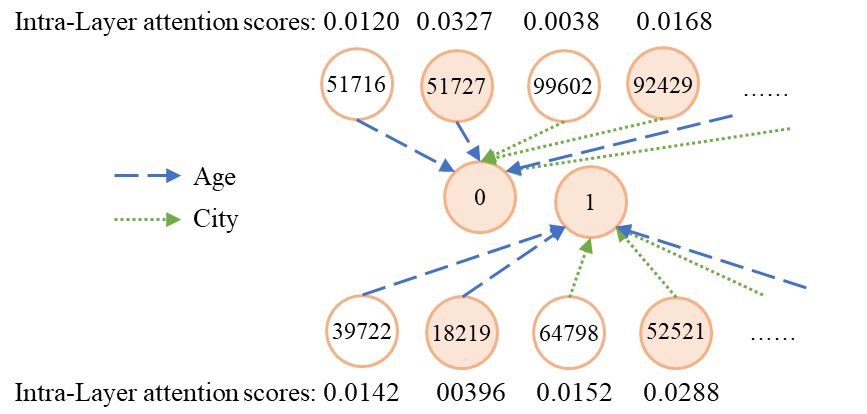}
	\caption{An illustrative example of two users ($uid=0$ and $uid=1$) from Data3, which is a loan scenario. \textit{Age} and \textit{City} are used to construct the multiplex graph. Circles in dark are positive samples, i.e., defaulted users. We can see that the intra-layer attention scores of defaulted neighbors are larger.}
	\vspace{-10pt}
	\label{fig-case}
\end{figure}

\subsubsection{Case study.}
Finally, we show a case study from Data3, where we need to predict where a loan user will be defaulted. The results are shown in Figure \ref{fig-case}, from which we can see that the intra-layer attention scores between defaulted users (in dark) are larger than others. It verifies that the sample relations can be an indicator of final prediction.

\subsection{The Computational Overhead}
In this section, we evaluate the computational overhead of the proposed TabGNN. For simplicity, we directly report the average time cost of running AutoFE with and without TabGNN on the largest classification and regression datasets, respectively, in Table \ref{table-dataset}, i.e., JD for classification, and Data8 for regression. We can see that the increased computational overhead is around two times. Considering TabGNN is a neural network based model and the performance gains in Table \ref{table-performance}, the computational is acceptable. Moreover, since we provide an easy-to-plug method to integrate TabGNN with AutoFE, in practice, customs can decide whether to choose TabGNN based on their computational budgets.

\begin{table}[h]
	\setlength\tabcolsep{6pt}
	\centering
	\caption{The training cost in hours of AutoFE with and without TabGNN on JD and Data8.}
	\vspace{-10pt}
	\begin{tabular}{ccc}
		\toprule
		 &AutoFE   & AutoFE (+TabGNN) \\
		\midrule
		JD (Classification)  &    $\sim$4h      &  $\sim$11h \\
		Data8 (Regression)  & $\sim$7h      &  $\sim$20h \\
		\bottomrule
	\end{tabular}
	\label{time-cost}
	\vspace{-10pt}
\end{table}

\section{Conclusion and Future work}
\label{sec-con}

In this paper, we propose a novel framework TabGNN for tabular data prediction, which is widely used in real-world prediction scenarios, such as fraud detection in commercial banks.
By explicitly and systematically modeling the sample relations, which are ignored by previous feature interaction methods for TDP, we propose a multiplex graph neural network to capture the multifaceted sample relations to improve the prediction performance.
Based on some practical heuristics, we construct a multiplex graph to model the
multifaceted sample relations.
Then for each sample, a multiplex graph neural network is designed to adaptively aggregate the representations of
neighbors from the multiplex graph.
Finally, to integrate TabGNN with our tabular solution AutoFE, we concatenate the original embeddings and the enhanced embeddings by TabGNN, which are then fed to AutoFE for final prediction.
Experiments on various datasets demonstrates that TabGNN can significantly improve the performance on all 11 datasets, including classification and regression tasks. It demonstrates the usefulness of sample relations for TDP scenarios.

For future work, we plan to deploy TabGNN in more scenarios from our business partners and explore model-based methods to construct the multiplex graphs beyond the current heuristic ones.

\section*{Acknowledgements}
This work was supported in part by The National Key Research and Development Program of China under grant 2018YFB1800804, the National Nature Science Foundation of China under U1936217,  61971267, 61972223, 61941117, 61861136003, Beijing Natural Science Foundation under L182038, Beijing National Research Center for Information Science and Technology under 20031887521, and research fund of Tsinghua University - Tencent Joint Laboratory for Internet Innovation Technology.


\vspace{-2pt}
\bibliographystyle{ACM-Reference-Format}
\bibliography{main}

\clearpage
\appendix

\section*{A. Specific preprocessing design for each dataset}
\label{append-jd-pre}

For JD dataset, the original data is a multi-table data set composed of 5 tables, including main table, user info, sku info, comments of skus, action records of users. The timestamp of the data is 2016-02-01 to 2016-04-15, We set the last 5 days as test sets. Since sku info and user info has no temporal constraint, We didn't do anything about them. For comments of skus and users' actions, we delete the records in in test sets. Most importantly, due to the difference of the task, we discarded the original main table and regenerated samples from users' actions.

Specifically, for users' action records, each record contains $user\_id$ and $sku\_id$. We generate a cartesian product of these two ids, name it $pair\_id$, group by $paid\_id$, construct a behavior sequence, and sort by time. For the processed action sequence, the time interval between each two adjacent records is calculated in the sequence of the sequence. If it exceeds one day (24 hours), it is cut at this position. After cutting, the original action sequence becomes sevaral samples (sequences). For each sequence, retrieve whether purchase action is included in its behavior sequence. If it contains, it is a positive sample (target=1), otherwise it is a negative sample (target=0). The time of the first action in the action sequence is taken as the time of the sample. So far we have completed the construction of a dataset including 5 tables for binary classification task.

Other data sets themselves can be applied to standard binary classification tasks, so there are no other special processing.

\section*{B. Feature transformation}

For all datasets, data types can be roughly divided into four types: categorical values, numerical values, text and timestamps. For text, we did not extract the semantic representation, but directly used it as a categorical value. For datasets with temporal constraints, the timestamp is treated as a numerical value. So far, all the features can be divided into two categories: categorical value and numerical value. It should be noted that there are no multi-value categorical features in all datasets.

\section*{C. Missing value}
Since there are usually many missing values in table data, the missing values need to be processed first. For categorical values, we replace the missing value with a new value, and for numerical values, we use 0 to fill in and normalize the values of each dimension.


\section*{D. Feature Encoders}
\label{append-feature-encoder}
In encode part, the encoder takes preprocessed data as input and outputs a fixed-length vector representation. In input data, there are categorical features $x_{cat}=\{x_{cat1}, x_{cat2}, \dots , x_{catM}\}$ and numerical features  $x_{num}=\{x_{num1}, x_{num2}, \dots , x_{numN}\}$. For categorical features, each feature $x_{cati}$ is a one-hot vector with high dimension, we transform it into a low-dimension dense vector $e_{cati}=V^{i}x_{cati}$. Then, all features are concated
\begin{equation}
e = [e_{cat1}, e_{cat2},\dots, e_{catM}, x_{num1}, x_{num2}, \dots, x_{numN}]
\end{equation}

Then through MLP, a fixed-length embedding is obtained. Except for the first layer, the input and output dimensions of each layer of MLP are consistent with the length of final embedding.

\section*{E. Hyper-parameter Tuning}
\label{append-hyper-tuning}

For hyper-parameter tuning, we use Hyperopt\footnote{https://github.com/hyperopt/hyperopt} and the stop criterion is a maximum of 100 trial or 48 hours for running time. For all experiments, we use Adam as the optimizer.
Hyper-parameter search range and best hyper-parameter are shown in Table \ref{hyper-parameter-search-gnn}, where $hidden\_dim$ is the dimension of embeddings after encoding (also the dimension of node embedding in graph) and $layer\_size$ is the layers of MLP in encoder.

\begin{table}[h]
	\setlength\tabcolsep{3pt}
	\centering
	\caption{Hyper-parameter search range.}
	\vspace{-5pt}
	\begin{tabular}{cc}
		\toprule
		Hyper-parameters  & Search range \\ \midrule
		$lr$              & $[10^{-6},10^{-3}]$ \\
		$dropout$         & $[0,1]$ \\
		$hidden\_dim$     & $\{64,128,256\}$ \\
		$weight\_decay$   & $[0,1]$ \\
		$attention\_head$ & $\{2,4\}$ \\
		$layer\_size$     & $\{1,2,3,4\}$ \\
		\bottomrule
	\end{tabular}
	\label{hyper-parameter-search-gnn}
\end{table}

\end{document}